\documentclass{article}
\PassOptionsToPackage{numbers,compress}{natbib}

\usepackage[dblblindworkshop, final]{neurips_ACA_2025}

\workshoptitle{\\ Workshop on Algorithmic Collective Action (ACA@NeurIPS 2025)}

\usepackage[utf8]{inputenc} 
\usepackage[T1]{fontenc}    
\usepackage{graphicx} 
\usepackage{hyperref}       
\usepackage{url}            
\usepackage{booktabs}       
\usepackage{amsfonts}       
\usepackage{nicefrac}       
\usepackage{microtype}      
\usepackage{xcolor}         

\usepackage{booktabs}

\usepackage{tikz}
\usetikzlibrary{positioning,fit,arrows.meta,shapes,calc}

\title{Collective Narrative Grounding: Community-Coordinated Data Contributions to Improve Local AI Systems}

\author{%
  Zihan Gao\thanks{Co-first Author.} \\
  Information Science\\
  University of Wisconsin-Madison\\
  Madison, WI \\
  \texttt{zihan.gao@wisc.edu} \\
  \And
  Mohsin Y. K. Yousufi\footnotemark[1] \\
  Digital Media \\
  Georgia Tech \\
  Atlanta, GA \\
  \texttt{yousufi@gatech.edu} \\
  \And
  Jacob Thebault-Spieker \\
  Information Science\\
  University of Wisconsin-Madison\\
  Madison, WI \\
  \texttt{jacob.thebaultspieker@wisc.edu} \\
}

\begin{document}

\maketitle




\begin{abstract}
Large language model (LLM) question-answering systems often fail on community-specific queries, creating ``knowledge blind spots'' that marginalize local voices and reinforce epistemic injustice. We present \textbf{Collective Narrative Grounding}, a participatory protocol that transforms community stories into structured \emph{narrative units} and integrates them into AI systems under community governance. Learning from three participatory mapping workshops with $N{=}24$ community members, we designed elicitation methods and a schema that retain narrative richness while enabling entity, time, and place extraction, validation, and provenance control. To scope the problem, we audit a county-level benchmark of 14{,}782 local information QA pairs, where factual gaps, cultural misunderstandings, geographic confusions, and temporal misalignments account for 76.7\% of errors. On a participatory QA set derived from our workshops, a state-of-the-art LLM answered fewer than 21\% of questions correctly without added context, underscoring the need for local grounding. The missing facts often appear in the collected narratives, suggesting a direct path to closing the dominant error modes for narrative items. Beyond the protocol and pilot, we articulate key \emph{design tensions}, such as representation and power, governance and control, and privacy and consent, providing concrete requirements for retrieval-first, provenance-visible, locally governed QA systems. Together, our taxonomy, protocol, and participatory evaluation offer a rigorous foundation for building community-grounded AI that better answers local questions.
\end{abstract}

\begin{figure}[t]
    \centering
    \includegraphics[width=0.9\linewidth]{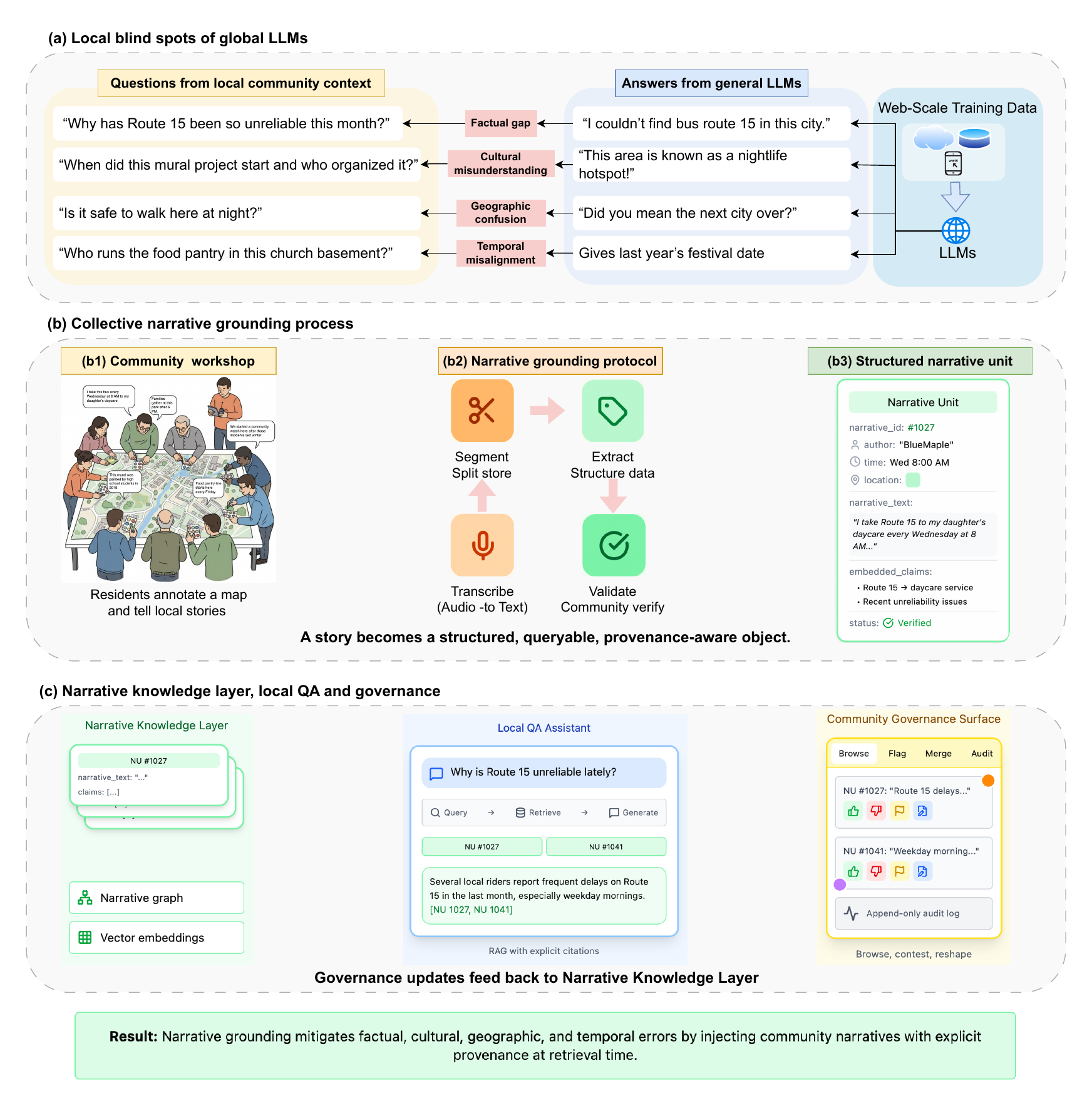}
    \caption{\textbf{Collective Narrative Grounding Framework.} (a) Global LLMs exhibit systematic local knowledge gaps across factual, cultural, geographic, and temporal dimensions. (b) Community members share place-based stories through participatory mapping; stories are transformed into structured, validated narrative units. (c) Narrative units form a queryable knowledge layer that powers a retrieval-augmented local QA system under community governance.}
    \label{fig:narrative_grounding_framework}
\end{figure}

\section{Introduction}
\label{sec:introduction}

Global question-answering systems powered by large language models have achieved remarkable breadth of knowledge, yet they often struggle with \emph{local} knowledge: information tied to specific communities, places, or cultures \cite{gao2025aturing,zhu2024quite,jang2024place,gao2024journeying}. For example, an LLM-based assistant might falter when asked about a neighborhood’s historical event or a local festival known only to residents. This deficiency is not just a technical gap; it reflects whose knowledge has been prioritized in training data and whose has been omitted. Researchers have framed such gaps as issues of \emph{epistemic injustice}, where the knowledge of marginalized or peripheral communities is undervalued or excluded \cite{yousufi2025credibility}. Missing local context can also create \emph{information gaps} that underrepresent rural or low-income communities~\cite{hickman_understanding_2021, thebault-spieker_distance_2018, thebault-spieker_geographic_2018} and leave space for misinformation to spread \cite{flores2022datavoidant,greene2024evaluation,aubin2024not}. Addressing the lack of local knowledge in AI systems is thus important for both improving utility and ensuring fairness in information access.

In this paper, we introduce a \textbf{Narrative Grounding Protocol} for augmenting QA systems with community-sourced narratives (See Figure~\ref{fig:narrative_grounding_framework}). The protocol specifies how to collect, process, and integrate locality-specific stories into an AI system’s knowledge pipeline. Our approach emerges from participatory workshops with 24 community members as co-designers, who shared local histories and practices and co-imagined how an assistant should use these narratives to answer questions.
Motivating this intervention, we first audit failures on a county-level local-knowledge benchmark and find that four categories, such as factual gaps, cultural misunderstandings, geographic confusion, and temporal, account for 76.7\% of errors, i.e., failure modes that locally grounded narratives can directly address. We then test this pathway on a small participatory QA set derived from the workshop: a state-of-the-art LLM answers fewer than 21\% of questions correctly without added context, often responding with generic or erroneous content. Taken together, these results clarify both the problem (systematic local information gaps) and the remedy (community narratives as targeted grounding), complementing broader evidence of LLM blind spots in underrepresented domains \cite{moayeri2024worldbench}.

Through \textit{Narrative Grounding}, we contribute: \textbf{(i)} by quantifying \emph{local knowledge blind spots} via a failure taxonomy, showing that the majority of errors are directly addressable with locally grounded context. \textbf{(ii)}  the \emph{Narrative Grounding Protocol} for collecting and integrating community narratives into QA through a participatory design workshop with community members, from data capture to retrieval-time integration and provenance filtering. \textbf{(iii)} We distill design implications and data-governance considerations for building equitable, community-informed AI systems.

\section{Background and Related Work}
\label{sec:background}

\subsection{Local Knowledge and Information Gaps in AI}

Our work relates to research on the limitations of AI knowledge bases regarding local or community-specific information. ``Data voids'' refer to topics or locales with little high-quality data available, which can be exploited by low-quality or malicious information\cite{flores2022datavoidant,greene2024evaluation,aubin2024not}. \citet{flores2022datavoidant} introduced the concept in the context of political information on social media and developed an AI system to proactively fill such voids. For the place representations of LLMs, recent evaluations have formalized the issue: WorldBench benchmark quantifies geographic disparities in LLM knowledge, finding that even cutting-edge models have systematically lower factual recall for questions about less-documented regions \cite{moayeri2024worldbench}. Similarly, \citet{lan2025benchmarking} benchmarked LLMs on \emph{local life services} questions and found performance lags compared to general knowledge queries. These studies align with older findings from Q\&A platforms. For example, \citet{park2014understanding} observed that users often sought ``here-and-now'' local information, and that achieving \textit{localness} of answers required contributions from those with on-the-ground knowledge. More broadly, scholars like \citet{thebault-spieker_distance_2018} and \citet{johnson2016not} have documented systemic patterns that underrepresent rural places in important information resources that get used as training data for AI tools. \citet{dudy2025unequal} showed that popular LLMs’ recommendations exhibited geographical bias, favoring well-known locations and neglecting peripheral ones. These gaps underline a need for \emph{community-sourced knowledge}. Our approach to fill this need is grounded in community participation and narrative data, distinguishing it from purely algorithmic solutions like model fine-tuning on existing (often sparse) local texts.

\subsection{Integrating Local Knowledge into AI Systems}
Technically, incorporating local knowledge into AI models can take multiple forms. One approach is adapting LLMs on locally relevant corpora (e.g., local news archives, community forum posts). For example, \citet{han2024integrating} explored integrating region-specific knowledge into models to enhance public understanding of carbon neutrality policies. Likewise, \emph{localness} is multifaceted—beyond geography, it includes social and cultural familiarity \cite{gao2025aturing}. 
\citet{gao2025from} exemplify LLMs' local knowledge gaps by fusing social media videos about local places into a knowledge graph, then using that graph to inform an LLM’s responses. Their \textit{Localness-Aware LLM} concept shows promising gains in answering location-specific queries by accessing community-contributed content. These efforts are aligned with the broader trend of grounding LLMs in external data sources to improve factual accuracy and context relevance as well as for AI systems to provide not just correct answers but \emph{contextualized} ones that users find meaningful and trustworthy.


\subsection{Participatory Approaches and Narrative in HCI}

Integrating community knowledge into AI, however, poses significant design challenges, including effectively engaging the relevant communities and issues of scale. One approach for addressing the former is through \emph{participatory design} and storytelling. Community storytelling projects have demonstrated how narrative can capture rich local context: for example, \citet{cong2021collective} developed a context-aware platform for collective community storytelling, and \citet{halperin2023probing} built a conversational agent to document digital stories of housing insecurity with community members using ``narratives''. These systems show that narratives are a powerful medium for encoding local experiences. Moreover, involving community members in the design process can ensure the resulting technology aligns with local values and needs\cite{bala2024stories}. Narratives add cultural nuance often missing from factual datasets, but they raise tensions around informality, subjectivity, and ethics. Converting stories into machine-usable representations risks loss of meaning \cite{loukissasOpenDataSettings2021}, and user-generated narratives can be co-opted or misused without provenance and moderation safeguards \cite{starbird2019disinformation}. Inspired by these examples, we posit that local \emph{narratives}, such as personal stories, oral histories, and firsthand accounts, can serve as valuable data to ground QA systems in community-specific knowledge. We treat community stories as first-class data through a co-designed collection protocol and governance hooks, extending \emph{conscious data contribution} toward local QA \cite{vincent2021can}. Technically, this complements both targeted knowledge injection and retrieval-augmented pipelines for place-aware QA \cite{han2024integrating,gao2025from}.

\section{The Narrative Grounding Approach: Participatory grounding for narrative unit}
\label{sec:approach}

The Narrative Grounding protocol emerges from a series of in-situ participatory mapping workshops designed to elicit lived, place-based experiences from community members. Each session engaged participants from a single community Atlanta with a focus on hyper-local knowledge that is underrepresented online. Across \textit{3 workshops} we worked with \textit{[N=24]} residents \textit{(ages 18 - 55)} in a local university. 

\subsection{Protocol Design Principles}

During our workshops, we consistently encountered instances of frictions that are common in participatory workshops \cite{ledantecStrangersGateGaining2015} exhibited in the form of reluctance or hesitation to share stories. These frictions emerge from a mismatch between the goals, values, and expectations of the communities and the ``technocrats'' (e.g., researchers, platforms etc) \cite{Yousufi2023bridging}. As a result, the narrative grounding protocol deploys the following principles: 

\emph{Principle 1: Explicit Expert-Framing.} The participants are explicitly framed as the ``neighborhood expert'' and encouraged to engage with each other's stories and experiences. Instead of a series of pre-determined prompts, the facilitators are trained to allow for ambiguity, openness, and flexibility, and play a limited role. While the participants are constantly reminded of their position as the expert, inverting traditional power dynamics between facilitators and participants. 

\emph{Principle 2: Physical Scaffolding.} The core of the mapping workshop is a large-scale satellite projection of the neighborhood on a table covered with paper. At the start of each session, participants trace out the map and mark out their places of interest with markers. This initial low-stakes activity creates the initial geocoded anchor points for our narrative nodes and provides an accessible opening for the mapping. We utilize pre-collected data maps of the communities, including demographic distribution and income levels. The data projected on these maps can be switched on demand and is complementary to the discussion and stories being shared around the table. 

\emph{Principle 3: Asset-Based Framing.} We begin by asset-based questions to provide the opportunity for the community members to narrate the stories about their communities in their own way such as ``Where are your favorite spots in the neighborhood?'', instead of  deficit-based prompts, such as ``What are the issues in the area?'' 
\label{sec:ethics}

\emph{Principle 4: Ethical Engagement.} All workshop activities followed informed consent procedures. Participants agreed that their contributions could be used for research and publicly shared \emph{only} in de-identified form; no personal identifiers are stored, and examples are paraphrased where needed to prevent re-identification. Community validators could withdraw or redact content at any time.

\subsection{Narrative Elicitation Methods}

To further elicit the stories from the community, our workshops focused on assisting participants to come up with place-based, hyper-local, and specific narratives. These narratives emerge through story-based prompting, where facilitators ask participants to share stories or experiences about specific events. A facilitator might ask ``Tell me about the last time you visited this area, and what did you experience?'' instead of ``How's the public transit in the area?''. These prompts aim to elicit details such as specific time-, place-, and body-based experiences. 

Participants are then encouraged to build on one another's contributions. In our workshops, we found that this was a powerful method of facilitating an active conversation. Participants shared their own experience of, for instance, public safety, affordability, accessibility, and negotiated claims made by one another. Some participants would even provide new knowledge about community initiatives happening in the area, or histories about the community and its temporal life. This collective process aims to generate rich data and also illustrates how experiences connect to broader community patterns in rich, layered narratives. 

\subsection{From Stories to Structured Knowledge}

In order for the output of these workshops to be integrated into an algorithmic system, it is necessary to impose some computationally legible structure on these messy, qualitative oral stories. In other words, shifting these oral stories into the structured \verb|narrative unit| schema that we describe in Table \ref{tab:schema}. The narrative unit moves through the following steps: (i) \emph{Verbatim Transcription}: The audio/visual recordings from the workshops are transcribed verbatim. (ii) \emph{Human-in-the-Loop Segmentation}: Through NLP pipelines, each transcript is broken down into the individual narrative units. These narrative units are verified by both the researcher and community members. (iii) \emph{Schema Population}: Each narrative unit populates a structured schema (detailed below) with extracted entities, geocodes, and relationship markers. (iv) \emph{Validation}: The narrative units are then presented back to the community for validation and consent, making sure that they accurately represent the community's meaning.

For example, from the workshops one narrative would be: ``I come from [another university] and I tell people at [name of university] that I went there and they're like oh my God, you live downtown and I'm like its 10 minutes aways, like its realy not that bad .... so there's this idea that you can't go there after 8pm.'' This narrative embeds a claim, which then serves as the ``narrative unit'' for our system. 

Another example of the narrative would be: "After the festival in the park, it rained, it was all muddy, and I had to walk almost an hour [back home] in mud-covered boots without proper sidewalks." Each narrative unit is then operationalized as a data object based on the scheme below. 

\begin{table}[]
\centering
\small
\caption{The Narrative Unit Schema}
\begin{tabular}
{lp{0.7\linewidth}} 
\toprule
\textbf{Field} & \textbf{Description} \\
\midrule
\verb|narrative_id| & A unique identifier for the story. \\ 
\verb|author_pseudonym| & A persistent but anonymized identifier for the contributor. \\ 
\verb|timestamp| & The timestamp of when the event occurred or when the narrative was submitted. \\ 
\verb|geocode| & GeoJSON data specifying a point, line, or polygon. \\ 
\verb|narrative_text| & The full, unaltered text or transcript of the narrative. \\
\verb|embedded_claims[]| & An array of specific, factual statements extracted from the narrative. \\
\verb|media_links[]| & An array of URLs pointing to user-submitted photos, audio recordings, or supporting documents. \\
\verb|verification_status| & An overall status for the narrative (\verb|unverified|, \verb|community_verified|, \verb|disputed|, \verb|retracted|),  \\
\verb|community_flags[]| & An array of tags applied by community members for moderation and curation.\\ 
\verb|relationships[]| & An array of links to other narrative units, defining the edges of the knowledge graph.\\
\bottomrule
\end{tabular}
\label{tab:schema}
\end{table}


\section{Local Knowledge Gaps: From Failure Taxonomy to Participatory Grounding}

Building on the protocol in Section~\ref{sec:approach}, we first quantify the very gap it targets: we (i) construct a failure taxonomy on \textsc{LocalBench} to characterize where LLMs falter \cite{gao2026localbench} and (ii) run a small participatory QA check to see whether workshop narratives actually contain the missing facts for those failures. Together, these establish the target error modes and a plausibility test for narrative grounding.

We systematically evaluated errors on a county-level local-knowledge benchmark (\textsc{LocalBench}) comprising 14{,}782 validated QA pairs from 526 U.S.\ counties drawn from complementary sources, such as structured county indicators (e.g., ACS/USDA/NRHP), local news, and online discussion threads. Questions span narrative (non-numerical) and numerical tasks aligned to a Localness framework (physical, cognitive, relational domains). Models are run in closed-book and web-augmented settings. and assessed using exact/semantic matches for non-numerical answers, a strict numeric-accuracy rule for numerical answers, and a LLM-judge-based binary correctness metric; we also track answer rate to capture selective refusal. This setup allows us to separate knowledge gaps from reasoning/calibration failures in local QA.

We then audited 1{,}000 model failures on \textsc{LocalBench} across narrative (non-numerical) and numerical tasks spanning county indicators, local news, and community discourse, isolating knowledge gaps vs.\ reasoning/calibration issues.  The results show eight mutually exclusive categories. The top four: Factual Knowledge Gap (31.8\%), Cultural Misunderstanding (23.4\%), Geographic Confsion (12.4\%), and Temporal Misalignment (9.1\%), account for \textit{76.7\%} of all errors and are \emph{primarily addressable} via locally grounded narratives and better source selection.%
\footnote{Failures were stratified by task type, data source, and model; two trained judges labeled each case with consensus adjudication (raw agreement $=87\%$, Cohen's $\kappa = 0.852$).}

\begin{table}[ht]
\centering
\caption{Misinformation Vulnerability Analysis. Model failures create systematic opportunities for false narratives that participatory mapping can address.}
\begin{tabular}{lcc}
\hline
\textbf{Error Category} & \textbf{Frequency} & \textbf{Misinformation Exploit}  \\
\hline
Knowledge Gaps & 67.3\% & False claims fill information voids  \\
Cultural Disconnect & 23.4\% & Exploits authenticity deficits  \\
Temporal Lag & 15.8\% & Spreads outdated narratives  \\
Source Absence & 12.1\% & Creates "data void" conditions \\
\hline
\end{tabular}
\label{tab:misinformation_vulnerability}
\end{table}

The taxonomy points directly to a remedy: community-authored, locality-specific narratives should reduce \emph{factual, cultural, temporal} and \emph{source} errors (and improve answer rate). 
To test this remedy pathway, we ran an exploratory baseline on a small, participatory QA set derived from the participatory workshops. We used OpenAI's GPT-5 via API as a representative model and sampled 20 factual/descriptive questions likely to admit objective answers. Annotators were (i) a \emph{community validator} with local knowledge and (ii) a \emph{research annotator}, labeling independently with adjudication on disagreement (raw agreement $=84.2\%$, Cohen's $\kappa = 0.812$). The results show that 4/20 answers were fully correct; 12/20 were partially correct or vague (missed locally salient specifics); 3/20 were incorrect/hallucinated. These incorract QAs are typically misidentifying local officials, conflating nearby places, or giving outdated event details.

The baseline aligns with the taxonomy: most errors are factual, cultural, temporal, or source-related. In a majority of non-correct cases, the missing facts were already present in the workshop narratives, indicating that supplying this material should close the gap. Using narrative grounding as retrieval-first context with hyper-local provenance filtering, we can therefore expect (i) higher answer rates, (ii) incorrect$\to$partial and partial$\to$correct conversions on narrative items, and (iii) reduced retrieval hallucination.

\section{Narrative Grounding: A system for Constructive Algorithmic Collective Action}

\begin{figure}[t]
    \centering
    \includegraphics[width=\linewidth]{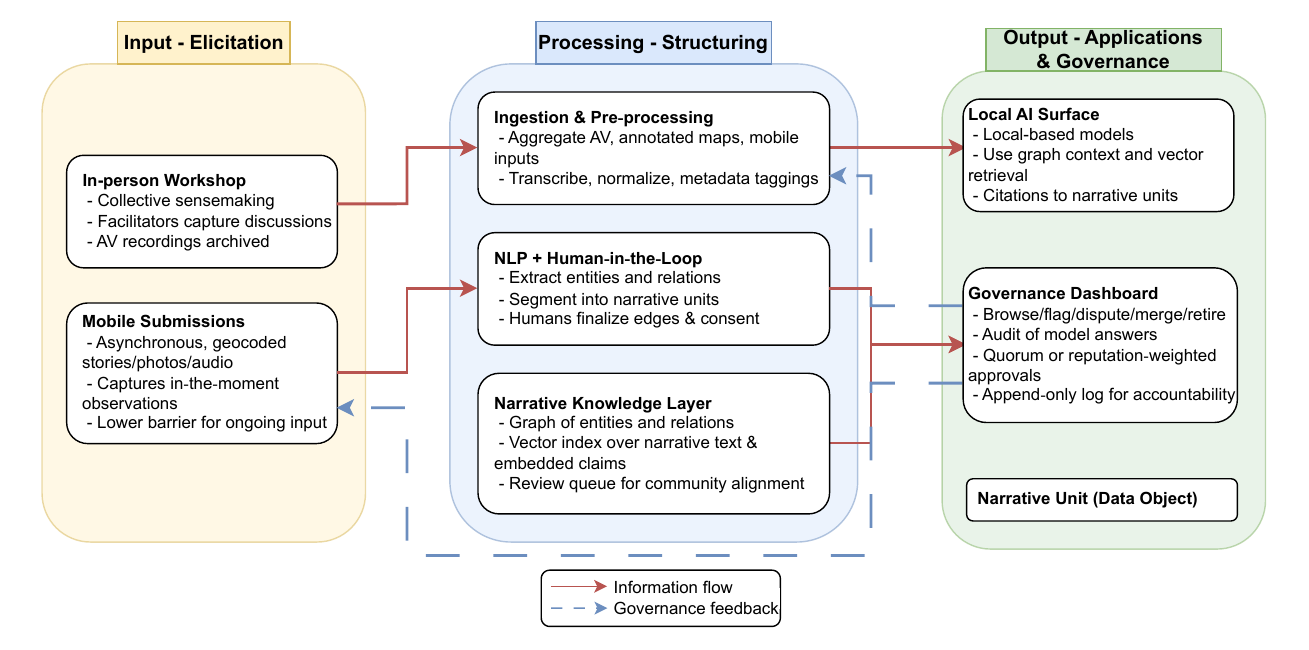}
    \caption{\textbf{System architecture of the Collective Narrative Grounding Framework.} The system combines community input and human-in-the-loop processing to build a structured narrative knowledge layer supporting local question answering and accountable governance. Input modules collect local stories; processing modules structure them into validated narrative units; output modules power local AI applications and community review.}
    \label{fig:system}
\end{figure}

Our system is designed as a continuous loop between ongoing community governance, and community knowledge, which is then deployed for local AI. The system consists of three components: Elicitation, Structuring and Governance (See Figure~\ref{fig:system}). 

\subsection{System architecture}
\emph{Input (Elicitation)}: The core of the system is a hybrid elicitation protocol designed for maximum inclusivity. In-person participatory mapping workshops, are captured through audio/video (AV) recordings, creating a rich archive of the community discussions. These workshops are complemented by mobile-first tools that allow for asynchronous, geocoded submission of stories, photos, and audio clips. This dual approach allows the system to capture both deeply considered reflections from facilitated discussions and immediate, in-the-moment observations from daily life.

\emph{Processing (Structuring)}: Raw materials (AV, annotated maps, mobile submissions) are transformed into a \emph{narrative knowledge layer}: (i) a graph of \emph{narrative units} linked by relations (corroborates/disputes/extends/near-in-space/near-in-time), and (ii) a vector index over \verb|narrative_text| and \verb|embedded_claims[]| for retrieval. NLP pipelines assist with entity/time/place extraction and claim detection, and humans finalize segmentation and relationship edges. Each unit is queued for community review in follow-up sessions to align structure with local understanding and consent.


\emph{Output (Application \& Governance)}: The structured data representation of these narrative units enable two types of user-facing computational systems (1) \emph{Local AI surface:} Retrieval-Augmented Generation (RAG) uses a vector index over narrative units, together with graph context, to retrieve provenance-tracked evidence and condition LLM responses with inline citations; the same corpus can also support lightweight fine-tuning of specialized local models. (2) \emph{Community governance surface:} A dashboard enables community members/curators to browse, flag, dispute, merge, or retire narratives; view and understand the provenance of narratives; and audit model answers that cited community content. Lightweight, DAO-inspired consensus could be feasible as well (e.g., quorum-based approval or reputation-weighted votes for sensitive changes), recorded in an append-only log for accountability.

\subsection{Design Tensions}
Grounding QA in community narratives raises persistent tensions that look different across contexts (rural vs.\ urban, multi-ethnic coalitions, Indigenous sovereignty, immigrant/diaspora groups, youth/elder forums). We frame them to guide algorithmic collective action (ACA) infrastructures.

\textbf{Representation \& Power.}
Whose stories become legible to the system? Narrative corpora risk amplifying the voices of those with time, access, or institutional backing, while quieter or minoritized perspectives remain under-documented. Tensions arise between making a coherent ``local profile'' and preserving plurality, dissent, and internal contradictions; between ``speaking with'' and ``speaking for''; and between community self-description and externally curated accounts.


\textbf{Governance \& Control.}
Ownership and decision rights over the corpus, models, and release practices are contested. Communities are not monolithic, and legitimacy can be challenged by factions with different values, priorities, or risk tolerances \cite{hardy_turn_2024}. Frictions emerge around who sets inclusion rules, who adjudicates disputes, how authority shifts over time, and where boundaries of ``the community'' begin and end, particularly for cross-neighborhood, cross-jurisdiction, or diasporic settings. A more comprehensive approach requires protocols for coordination before, during, and after workshops, with clear mechanisms for resolving conflicting narratives. It must also minimize added sociotechnical burden on communities already working to build equitable, accessible information infrastructures.
 

\textbf{Privacy, Consent \& Agency.}
Narratives often carry sensitive, identifiable, or context-dependent information. A core tension is between de-identification and utility: the more specific the story, the more useful --- and the more re-identifiable. Consent is also temporal and situational; what felt shareable in a workshop may feel risky once indexed, retrieved, and recontextualized by models, especially for groups facing surveillance or stigma.  Focusing on this tension, and the trade-offs therein --- between the potential value of the technology itself, and the risks associated with more datafication of people's lives --- remains an open area of study. While community-driven governance models move the locus of power from centralized platforms and corporations, this structure may still contribute to similar power asymmetries on a micro-scale. Any effort to bring about a truly collective algorithmic system must distribute this power down to its last member.

{\small
\bibliographystyle{abbrvnat}
\bibliography{main}

\begin{thebibliography}{28}
\providecommand{\natexlab}[1]{#1}
\providecommand{\url}[1]{\texttt{#1}}
\expandafter\ifx\csname urlstyle\endcsname\relax
  \providecommand{\doi}[1]{doi: #1}\else
  \providecommand{\doi}{doi: \begingroup \urlstyle{rm}\Url}\fi

\bibitem[Aubin Le~Qu{\'e}r{\'e} et~al.(2024)Aubin Le~Qu{\'e}r{\'e}, Naaman, and Fields]{aubin2024not}
M.~Aubin Le~Qu{\'e}r{\'e}, M.~Naaman, and J.~Fields.
\newblock Not quite filling the void: Comparing the perceptions of local online groups and local media pages on facebook.
\newblock \emph{Proceedings of the ACM on Human-Computer Interaction}, 8\penalty0 (CSCW1):\penalty0 1--22, 2024.

\bibitem[Bala et~al.(2024)Bala, Nisi, and Nunes]{bala2024stories}
P.~Bala, V.~Nisi, and N.~J. Nunes.
\newblock Stories as boundary objects: Digital storytelling with migrant communities for heritage discourses.
\newblock \emph{Proceedings of the ACM on Human-Computer Interaction}, 8\penalty0 (CSCW1):\penalty0 1--32, 2024.

\bibitem[Cong et~al.(2021)Cong, Cheng, Zhang, and Louie]{cong2021collective}
N.~Cong, K.~Cheng, H.~Zhang, and R.~Louie.
\newblock Collective narrative: Scaffolding community storytelling through context-awareness.
\newblock In \emph{Companion Publication of the 2021 Conference on Computer Supported Cooperative Work and Social Computing}, pages 40--43, 2021.

\bibitem[Dudy et~al.(2025)Dudy, Tholeti, Ramachandranpillai, Ali, Li, and Baeza-Yates]{dudy2025unequal}
S.~Dudy, T.~Tholeti, R.~Ramachandranpillai, M.~Ali, T.~J.-J. Li, and R.~Baeza-Yates.
\newblock Unequal opportunities: Examining the bias in geographical recommendations by large language models.
\newblock In \emph{Proceedings of the 30th International Conference on Intelligent User Interfaces}, pages 1499--1516, 2025.

\bibitem[Flores-Saviaga et~al.(2022)Flores-Saviaga, Feng, and Savage]{flores2022datavoidant}
C.~Flores-Saviaga, S.~Feng, and S.~Savage.
\newblock Datavoidant: An ai system for addressing political data voids on social media.
\newblock \emph{Proceedings of the ACM on human-computer interaction}, 6\penalty0 (CSCW2):\penalty0 1--29, 2022.

\bibitem[Gao et~al.(2024)Gao, Cranshaw, and Thebault-Spieker]{gao2024journeying}
Z.~Gao, J.~Cranshaw, and J.~Thebault-Spieker.
\newblock Journeying through sense of place with mental maps: characterizing changing spatial understanding and sense of place during migration for work.
\newblock \emph{Proceedings of the ACM on Human-Computer Interaction}, 8\penalty0 (CSCW2):\penalty0 1--31, 2024.

\bibitem[Gao et~al.(2025{\natexlab{a}})Gao, Justin, and Jacob]{gao2025aturing}
Z.~Gao, C.~Justin, and T.-S. Jacob.
\newblock A turing test for ''localness'': Conceptualizing, defining, and recognizing localness in people and machines.
\newblock \emph{arXiv preprint arXiv:2505.07282}, 2025{\natexlab{a}}.

\bibitem[Gao et~al.(2025{\natexlab{b}})Gao, Liu, Xu, and Thebault-Spieker]{gao2025from}
Z.~Gao, J.~L. Liu, Y.~Xu, and J.~Thebault-Spieker.
\newblock From clips to communities: Fusing social video into knowledge graphs for localness-aware llms.
\newblock In \emph{Companion of the Computer-Supported Cooperative Work and Social Computing (CSCW Companion ’25)}, page~8, New York, NY, USA, October 18--22 2025{\natexlab{b}}. Association for Computing Machinery.
\newblock \doi{10.1145/3715070.3749277}.
\newblock URL \url{https://doi.org/10.1145/3715070.3749277}.

\bibitem[Gao et~al.(2026)Gao, Xu, and Thebault-Spieker]{gao2026localbench}
Z.~Gao, Y.~Xu, and J.~Thebault-Spieker.
\newblock Localbench: Benchmarking llms on county-level local knowledge and reasoning.
\newblock In \emph{Proceedings of the 40th Annual AAAI Conference on Artificial Intelligence (AAAI 2026)}, 2026.

\bibitem[Greene et~al.(2024)Greene, Pisharody, Guevara, Evans, and Shapiro]{greene2024evaluation}
K.~T. Greene, N.~Pisharody, A.~Guevara, N.~Evans, and J.~N. Shapiro.
\newblock An evaluation of online information acquisition in us news deserts.
\newblock \emph{Scientific reports}, 14\penalty0 (1):\penalty0 27780, 2024.

\bibitem[Halperin et~al.(2023)Halperin, Hsieh, McElroy, Pierce, and Rosner]{halperin2023probing}
B.~A. Halperin, G.~Hsieh, E.~McElroy, J.~Pierce, and D.~K. Rosner.
\newblock Probing a community-based conversational storytelling agent to document digital stories of housing insecurity.
\newblock In \emph{Proceedings of the 2023 CHI Conference on Human Factors in Computing Systems}, pages 1--18, 2023.

\bibitem[Han et~al.(2024)Han, Cong, Yu, Tang, and Wei]{han2024integrating}
T.~Han, R.-G. Cong, B.~Yu, B.~Tang, and Y.-M. Wei.
\newblock Integrating local knowledge with chatgpt-like large-scale language models for enhanced societal comprehension of carbon neutrality.
\newblock \emph{Energy and AI}, 18:\penalty0 100440, 2024.

\bibitem[Hardy and Thebault-Spieker(2024)]{hardy_turn_2024}
J.~Hardy and J.~Thebault-Spieker.
\newblock A {Turn} to {Assets} in {Community}-{Based} {Computing} {Research}: {Tradeoffs}, {Deficits}, and {Neoliberalism} in {Technological} {Development}.
\newblock \emph{Proc. ACM Hum.-Comput. Interact.}, 8\penalty0 (CSCW1):\penalty0 14:1--14:20, Apr. 2024.
\newblock \doi{10.1145/3637291}.
\newblock URL \url{https://dl.acm.org/doi/10.1145/3637291}.

\bibitem[Hickman et~al.(2021)Hickman, Pasad, Sanghavi, Thebault-Spieker, and Lee]{hickman_understanding_2021}
M.~G. Hickman, V.~Pasad, H.~K. Sanghavi, J.~Thebault-Spieker, and S.~W. Lee.
\newblock Understanding {Wikipedia} {Practices} {Through} {Hindi}, {Urdu}, and {English} {Takes} on an {Evolving} {Regional} {Conflict}.
\newblock \emph{Proceedings of the ACM on Human-Computer Interaction}, 5\penalty0 (CSCW1):\penalty0 34:1--34:31, Apr. 2021.
\newblock \doi{10.1145/3449108}.
\newblock URL \url{https://dl.acm.org/doi/10.1145/3449108}.

\bibitem[Jang et~al.(2024)Jang, Chen, Kang, Kim, Lee, Duarte, and Ratti]{jang2024place}
K.~M. Jang, J.~Chen, Y.~Kang, J.~Kim, J.~Lee, F.~Duarte, and C.~Ratti.
\newblock Place identity: a generative ai’s perspective.
\newblock \emph{Humanities and Social Sciences Communications}, 11\penalty0 (1):\penalty0 1--16, 2024.

\bibitem[Johnson et~al.(2016)Johnson, Lin, Li, Hall, Halfaker, Sch{\"o}ning, and Hecht]{johnson2016not}
I.~L. Johnson, Y.~Lin, T.~J.-J. Li, A.~Hall, A.~Halfaker, J.~Sch{\"o}ning, and B.~Hecht.
\newblock Not at home on the range: Peer production and the urban/rural divide.
\newblock In \emph{Proceedings of the 2016 CHI conference on Human Factors in Computing Systems}, pages 13--25, 2016.

\bibitem[Lan et~al.(2025)Lan, Feng, Lei, Shi, and Li]{lan2025benchmarking}
X.~Lan, J.~Feng, J.~Lei, X.~Shi, and Y.~Li.
\newblock Benchmarking and advancing large language models for local life services.
\newblock In \emph{Proceedings of the 31st ACM SIGKDD Conference on Knowledge Discovery and Data Mining V. 2}, pages 4566--4577, 2025.

\bibitem[Le~Dantec and Fox(2015)]{ledantecStrangersGateGaining2015}
C.~A. Le~Dantec and S.~Fox.
\newblock Strangers at the {Gate}: {Gaining} {Access}, {Building} {Rapport}, and {Co}-{Constructing} {Community}-{Based} {Research}.
\newblock In \emph{Proceedings of the 18th {ACM} {Conference} on {Computer} {Supported} {Cooperative} {Work} \& {Social} {Computing}}, {CSCW} '15, pages 1348--1358, New York, NY, USA, Feb. 2015. Association for Computing Machinery.
\newblock ISBN 978-1-4503-2922-4.
\newblock \doi{10.1145/2675133.2675147}.
\newblock URL \url{https://dl.acm.org/doi/10.1145/2675133.2675147}.

\bibitem[Loukissas and Ntabathia(2021)]{loukissasOpenDataSettings2021}
Y.~A. Loukissas and J.~M. Ntabathia.
\newblock Open {Data} {Settings}: {A} {Conceptual} {Framework} {Explored} {Through} the {Map} {Room} {Project}.
\newblock \emph{Proc. ACM Hum.-Comput. Interact.}, 5\penalty0 (CSCW2):\penalty0 357:1--357:24, Oct. 2021.
\newblock \doi{10.1145/3479501}.
\newblock URL \url{https://doi.org/10.1145/3479501}.

\bibitem[Moayeri et~al.(2024)Moayeri, Tabassi, and Feizi]{moayeri2024worldbench}
M.~Moayeri, E.~Tabassi, and S.~Feizi.
\newblock Worldbench: Quantifying geographic disparities in llm factual recall.
\newblock In \emph{Proceedings of the 2024 ACM Conference on Fairness, Accountability, and Transparency}, pages 1211--1228, 2024.

\bibitem[Park et~al.(2014)Park, Kim, Lee, and Ackerman]{park2014understanding}
S.~Park, Y.~Kim, U.~Lee, and M.~Ackerman.
\newblock Understanding localness of knowledge sharing: a study of naver kin'here'.
\newblock In \emph{Proceedings of the 16th international conference on Human-computer interaction with mobile devices \& services}, pages 13--22, 2014.

\bibitem[Starbird et~al.(2019)Starbird, Arif, and Wilson]{starbird2019disinformation}
K.~Starbird, A.~Arif, and T.~Wilson.
\newblock Disinformation as collaborative work: Surfacing the participatory nature of strategic information operations.
\newblock \emph{Proceedings of the ACM on human-computer interaction}, 3\penalty0 (CSCW):\penalty0 1--26, 2019.

\bibitem[Thebault-Spieker et~al.(2018{\natexlab{a}})Thebault-Spieker, Halfaker, Terveen, and Hecht]{thebault-spieker_distance_2018}
J.~Thebault-Spieker, A.~Halfaker, L.~G. Terveen, and B.~Hecht.
\newblock Distance and {Attraction}: {Gravity} {Models} for {Geographic} {Content} {Production}.
\newblock In \emph{Proceedings of the 36th {Annual} {ACM} {Conference} on {Human} {Factors} in {Computing} {Systems}}, page~13, 2018{\natexlab{a}}.

\bibitem[Thebault-Spieker et~al.(2018{\natexlab{b}})Thebault-Spieker, Hecht, and Terveen]{thebault-spieker_geographic_2018}
J.~Thebault-Spieker, B.~Hecht, and L.~Terveen.
\newblock Geographic {Biases} {Are} '{Born}, {Not} {Made}': {Exploring} {Contributors}' {Spatiotemporal} {Behavior} in {OpenStreetMap}.
\newblock In \emph{Proceedings of the 2018 {ACM} {Conference} on {Supporting} {Groupwork}}, {GROUP} '18, pages 71--82, New York, NY, USA, 2018{\natexlab{b}}. ACM.
\newblock ISBN 978-1-4503-5562-9.
\newblock \doi{10.1145/3148330.3148350}.
\newblock URL \url{http://doi.acm.org/10.1145/3148330.3148350}.

\bibitem[Vincent and Hecht(2021)]{vincent2021can}
N.~Vincent and B.~Hecht.
\newblock Can" conscious data contribution" help users to exert" data leverage" against technology companies?
\newblock \emph{Proceedings of the ACM on Human-Computer Interaction}, 5\penalty0 (CSCW1):\penalty0 1--23, 2021.

\bibitem[Yousufi et~al.(2023)Yousufi, Loukissas, and Hyde]{Yousufi2023bridging}
M.~Yousufi, Y.~Loukissas, and A.~Hyde.
\newblock Bridging data and experiences: Engaging youth in digital civics through participatory mapmaking for resilience.
\newblock \emph{arXiv preprint arXiv.2309.16957}, 2023.

\bibitem[Yousufi et~al.(2025)Yousufi, Alexander, and Parvin]{yousufi2025credibility}
M.~Y.~K. Yousufi, C.~Alexander, and N.~Parvin.
\newblock Credibility boosters as a lens for understanding epistemic injustice in civic tech: The case of heat seek.
\newblock \emph{Proc. ACM Hum.-Comput. Interact.}, 9\penalty0 (7), Oct. 2025.
\newblock \doi{10.1145/3757573}.
\newblock URL \url{https://doi.org/10.1145/3757573}.

\bibitem[Zhu et~al.(2024)Zhu, Wang, and Liu]{zhu2024quite}
S.~Zhu, W.~Wang, and Y.~Liu.
\newblock Quite good, but not enough: Nationality bias in large language models--a case study of chatgpt.
\newblock \emph{arXiv preprint arXiv:2405.06996}, 2024.

\end{thebibliography}
}

\end{document}